\documentclass{ieeeaccess}
\usepackage{cite}
\usepackage{amsmath,amssymb,amsfonts}
\usepackage{algorithmic}
\usepackage{graphicx}
\usepackage{caption}
\usepackage{multirow}
\usepackage{textcomp}
\usepackage{stfloats}
\usepackage{hyperref}

\def\BibTeX{{\rm B\kern-.05em{\sc i\kern-.025em b}\kern-.08em
    T\kern-.1667em\lower.7ex\hbox{E}\kern-.125emX}}

\begin{document}
\history{None}
\doi{None}

\title{K-XLNet: A General Method for Combining Explicit Knowledge with Language Model Pretraining}

\author{\uppercase{Ruiqing Yan}\authorrefmark{1}, \uppercase{Lanchang Sun\authorrefmark{2}, Fang Wang\authorrefmark{1} and Xiaoming Zhang\authorrefmark{1}}}
\address[1]{College of Information Engineering, Beijing Institute of Petrochemical Technology, Beijing, 102617, CHN}
\address[2]{School of computer science, Beijing University of Posts and Telecommunications, Beijing, 100876, CHN}


\markboth
{Ruiqing Yan \headeretal: K-XLNet: A General Method for Combining Explicit Knowledge with Language Model Pretraining}
{Ruiqing Yan \headeretal: K-XLNet: A General Method for Combining Explicit Knowledge with Language Model Pretraining}

\corresp{Corresponding author: Fang Wang (e-mail: fangwang@bipt.edu.cn)}

\begin{abstract}
Though pre-trained language models such as Bert and XLNet, have rapidly advanced the state-of-the-art on many NLP tasks, they implicit semantics only relying on surface information between words in corpus. Intuitively, background knowledge influences the efficacy of understanding. Inspired by this common sense, we focus on improving model pretraining by leveraging explicit knowledge. Different from recent research that optimize pretraining model by knowledge masking strategies, we propose a simple but general method to combine explicit knowledge with pretraining. To be specific, we first match knowledge facts from knowledge graph (KG) and then add a knowledge injunction layer to transformer directly without changing its architecture. The present study seeks to find the direct impact of explicit knowledge on transformer per-training. We conduct experiments on various datasets for different downstream tasks. The experimental results show that solely by adding external knowledge to transformer can improve the learning performance on many NLP tasks.
  
\end{abstract}


\titlepgskip=-15pt

\maketitle

\section{Introduction}
\label{sec:introduction}


\PARstart{R}{ecently}, substantial work has shown that pre-trained models\cite{peters-etal-2018-deep,Andrew15,GPT,devlin2018bert} can learn language representations over large-scale text corpora, which are beneficial for downstream NLP tasks. For example, XLNet\cite{yang2019xlnet} obtains the state-of-the-art results on twelve NLP tasks including reading comprehension, question answering and text classification. However, most existing works model the representations by predicting the missing word only through the contexts, without considering the background knowledge in the text. And trained on general-purpose large-scale text corpora, the models usually lack domain adaptability.   

Background knowledge influences the efficacy of understanding. It has been observed that one major step in improving reading is to improve prior knowledge of the topics being read\cite{reading1980}. The pre-trained transformers can model implicit semantic information between surface form words\cite{petroni2019}. But it is solely at the token level. Considering Background knowledge can lead to better language understanding. For instance, given the sentence ``\textit{Xiaomi was officially listed on the main board of HKEx}'', the background knowledge includes \textit{Xiaomi} is a \textit{science and technology company}, \textit{HKEx} refers to \textit{Hong Kong Exchanges and Clearing Limited} and \textit{main board} is a \textit{economic term}. The explicit knowledge can help better understand the word sense and sentence topic.

Most recently, there are some improved models based on Bert\cite{ERNIE1,zhang2019ernie,liu2019k} or GPT\cite{rosset2020knowledgeaware}, which prove that injecting extra knowledge information can significantly enhance original models. Differently, this paper seeks to find the direct impact of explicit knowledge on transformer pretraining. Based on XLNet, we propose a simple but general method for combining knowledge without changing the model architecture. Given a sentence, We first use a simple dictionary look-up method to map its background knowledge. A knowledge injection layer is designed to combine the background knowledge with the original sentence, in a way that is close to natural language and accepted by XLNet without losing the structure information. Finally, we take the output of the knowledge injection layer directly as the input for XLNet. A three-stage model training method is proposed to save training time and hardware cost.To seek the impact of explicit knowledge, we leverage open domain and domain-specific knowledge to combine with XLNet and test their performances on various NLP tasks.

Our contributions in this paper are three folds: 1) proposal of a simple but general knowledge injection method for language model pretraining; 2) proposal of K-XLNet for an implementation of the proposed knowledge injection method on XLNet; 3) empirical verification of the effectiveness of K-XLNet on various downstream tasks.

\begin{figure*}[ht]
	\centering
	\includegraphics[width=15cm,height=9cm]{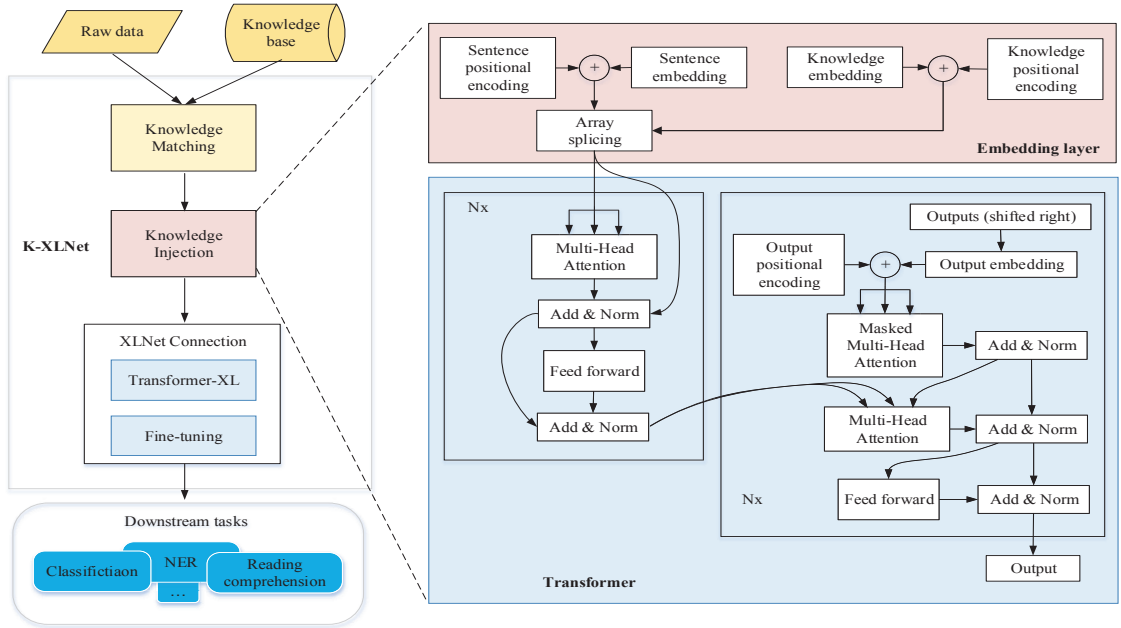}
	\caption{The overall framework of K-XLNet.}
	\label{fig:overall_framework}
\end{figure*}

The rest of the paper is organized as follows: Section \ref{sec:RelWork} summarizes related work.  In Section \ref{sec:Method}, we elaborate on our knowledge injection method taking XLNet as a running example. Section \ref{sec:Evaluation} reports experimental results, and finally Section \ref{sec:Conclusion} concludes the whole paper.

\section{RelatedWork}
\label{sec:RelWork}
In recent years, with the rapid development of deep learning, the pre-training technology\cite{Mikolov13,pennington2014glove,peters-etal-2018-deep,Andrew15,GPT} for the field 
of natural language processing has made great progress. Many efforts\cite{devlin2018bert,yang2019xlnet} are devoted to pre-training language representation models for learning language representations over large-scale corpus, and then utilizing the representations for downstream NLP tasks such as question answering\cite{zellers-etal-2018-swag}, text classification\cite{wang-etal-2018-glue}. For instance, XLNet\cite{yang2019xlnet} has obtained the state-of-the-art results on twelve NLP tasks.

Though the language representation models have achieved great success, they are still far from ready to serve as an ``unsupervised multitask learner.''\cite{rosset2020knowledgeaware} There are still gaps between model pretraining and task-specific fine-tuning\cite{lewis2020retrievalaugmented}. Pretraining models usually learn universal language representation from general-purpose large-scale text corpora, but lacking of domain or task specific knowledge. This leads to the need of huge effort for task-specific fine-tuning or over-parameterization\cite{kaplan2020scaling}. 

Recent work shows that combining with knowledge is a promising way to improve language model. Based on BERT and improved by refining the transformer architecture with knowledge, ERNIE\cite{ERNIE1}, ERNIE1(THU)\cite{zhang2019ernie} and K-Bert\cite{liu2019k} have revealed promising result in knowledge-driven applications such as named entity recognition, entity typing and relation extraction. Based on GTP2.0\cite{radford2019language}, KALM\cite{rosset2020knowledgeaware} significantly improves downstream tasks like zero-shot question-answering, by adding entity signals to the input of the transformer and an entity prediction task to the output.

This paper puts forward an attempt to combine knowledge with XLNet. Different with the above-mentioned methods, we focus on find the direct impact of explicit knowledge on transformer pretraining. Instead of changing the architecture of XLNet, we take sentences and their matched background knowledge facts as input into a knowledge injector composed of self-attention structure. The knowledge injector will combine the knowledge with original text and generate knowledge enriched output as the input of Transformer-XL. Further, we study the impact of various knowledge types on different downstream NLP tasks.

\section{Methodology}
\label{sec:Method}
For combing knowledge with XLNet, we propose a simple but general knowledge injection method called K-XLNet. Figure~\ref{fig:overall_framework} shows its overall framework. We can see that K-XLNet does not change the original architecture of XLNet. A knowledge injunction layer is designed and connected to Transformer-XL, so as to study the influence of background knowledge on model pretraining. We elaborate on K-XLNet in the following three subsections.

\begin{figure*}[h]
	\centering
	\includegraphics[width=17cm,height=5cm]{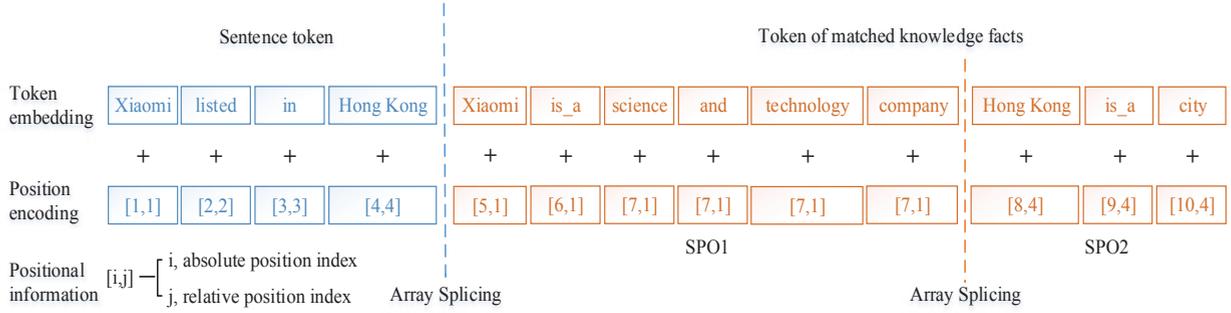}
	\caption{An example of embedding representation}
	\label{fig:embedding}
\end{figure*}

\subsection{Knowledge matching}
Regarding to the given text, how to effectively extract its background knowledge is a primary issue for knowledge combing with language models. We take the Subject-Predicate-Object (SPO) triples in knowledge graph (e.g. DBpedia) as the source of knowledge facts, and use a subject tokenizer to match the related knowledge facts for a given sentence. To be specific, the subject tokenizer segments the sentence into subjects using a surface form dictionary. It maps word n-grams to subjects in KG and then gets the related SPO triples as knowledge facts for the given sentence. For example, given the sentence ``\textit{Xiaomi listed in Hong Kong}'', we map two knowledge facts from KG as follows: (Xiaomi, is\_a, science and technology company), (Hong Kong, is\_a, city). Instead of using a more precise entity linker\cite{EL20}, we use a frequency-based dictionary look up to map the subject, because the dictionary look up is more efficient for large-scale text corpus and using a highly tuned entity linker may propagate its own biases into the transformer. 

\subsection{Knowledge Injection}
It is a challenge to fuse lexical, syntactic, and knowledge information in the pre-training procedure for language representation. Instead of designing a special pretraining objective for the fusion, we aim to integrate knowledge naturally into the original text in a way that conforms to the grammatical rules of natural language. For example, given the sentence ``\textit{Xiaomi listed in Hong Kong}'' and a knowledge fact (Xiaomi, is\_a, science and technology company), generate a knowledge-enriched sentence like ``\textit{Xiaomi, a science and technology company, listed in Hong Kong}''. By this way, we let the pretraining model to use clues in the knowledge-enriched text to learn word representation that better reflect how language conveys knowledge.

Therefore, we treat the knowledge injection problem like a machine translation problem and design a knowledge injector with the similar structure to the transformer used in the field of natural language translation, as the knowledge injunction layer shown in Fig.\ref{fig:overall_framework}. It mainly consists of two modules, i.e., embedding layer and transformer. For the input sentence and knowledge facts, embedding layer first converts them together with positional information into embedding representation and then feds into the transformer for knowledge combining. 

\subsubsection{Embedding layer}
The function of the embedding layer is to convert the given sentence and its related knowledge facts into an embedding representation that can be fed into the following transformer.

In order to express the positional information of the original sentence and the SPO triplet, we splice the original sentence with the matched knowledge triplets, When splicing the original sentence and SPO triples, we arrange the matching SPO triples according to the order of their corresponding words in the original sentence, such as
\begin{equation}\begin{array}{l}
		W_{1},W_{2},W_{3},W_{4},W_{5},W_{6}...SPO_{W_{2}},SPO_{W_{3}},SPO_{W_{5}}...
\end{array}\end{equation}

Next use a two-dimensional array composed of two elements to encode the position of each word. The first element is the sequence number of the word position, called absolute position index. The second element is the sequence number of the word that matches the subject in the original sentence, called relative position index. 

Fig.\ref{fig:embedding} shows an embedding example. The encoding of words in the original sentence is composed of two elements with the same value, because each word in the original sentence matches itself. For instance, the encode of the first word ``Xiaomi'' is $[1,1]$. For the matched knowledge triples, each SPO triple is horizontally spliced into a whole sequence, and S, P, O are sequentially indexed. For instance, the encode of the first matched knowledge subject ``Xiaomi'' is $[5,1]$. By this way, the information of the original sentence, matched SPO triples and their positions is all preserved. 

The calculation method of position code is to add absolute position index and relative position index and normalize them, that is, POS is equal to absolute position index and relative position index. From the calculation method of generating absolute position index and relative position index, it is easy to know that the value of absolute position index plus relative position index in different information positions is unique. After getting the position code, we use the following formula to normalize the position code.
\begin{equation}\begin{array}{l}
		\alpha=absolute position index\\
		\beta =relative position index\\
		pos=\alpha+\beta\\
		P E_{(p o s, 2 i)}=\sin \left(p o s / 10000^{2 i / d_{\text {model}}}\right) \\
		P E_{(p o s, 2 i+1)}=\cos \left(\text {pos} / 10000^{2 i / d_{\text {model}}}\right)
\end{array}\end{equation}

\subsubsection{Transformer}
We use a transduction model to inject the matched knowledge triples to the original sentence. Regarding the matched knowledge triples as a different language with the original sentence, we turn the knowledge injection problem into a machine translation problem, and translate them into a language that confirms to the natural language grammar. By this way, the knowledge-enriched output can be naturally used as the input of Transformer-XL. As shown in Fig.\ref{fig:overall_framework}, the transformer consists of an encoder and a decoder, which has the same structure with the typical Transformer\cite{NIPS_attention} used in machine translation area.  

\subsection{XLNet Connection}
XLNet is one of state-of-the-art natural language models. We take XLNet as a running example for combining knowledge with language model pre-training. The XLNet connected after the knowledge injection layer does not have a Tokenization module, namely the Transformer-XL. Tokenization and encoding have been performed in the embedding module.

\begin{figure}[h]
	\vspace{-6mm}
	\centering
	\includegraphics[width=8.5cm,height=5cm]{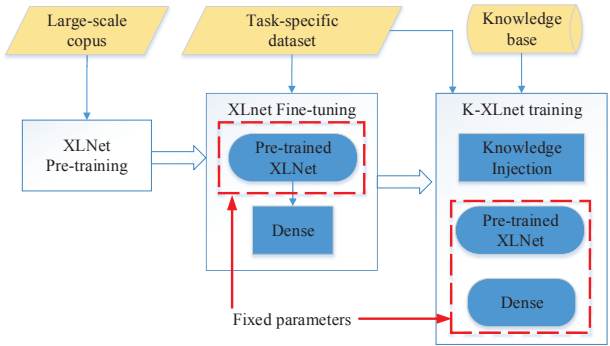}
	\caption{A simple method for training K-XLNet}
	\label{fig:kXLnettrain}
\end{figure}

Normally, retraining the pre-training model is necessary for model refinement by leverage knowledge  information. However, the cost of retraining is very high, both in terms of time and hardware cost. We propose a simple and general way to resolve this problem, inspired by the mainstream idea of pre-training and fine-tuning. Figure\ref{fig:kXLnettrain} shows the process of training K-XLNet. It mainly has three stages: XLNet pre-training, task-specific fine-tuning and K-XLNet training for the specific task. The first two stages are consistent with the usual two-stage pre-training model. Instead of pre-training K-XLNet on large-scale general corpus, we train it on specific tasks by leveraging external knowledge. In addition to cost savings, this approach makes it easy to flexibly test the effects of different knowledge bases on different downstream tasks. The following experiments show that this method is effective.

\section{Experiment}
\label{sec:Evaluation}
In this section, we evaluate the performance of K-XLNet through seven downstream tasks, among which one is an English task for a specific domain, six are Chinese tasks for open-domain. 

\subsection{Experiment Setup}

\subsubsection{Preprocessing}
We use the pre-trained word2vec\footnote{https://github\.com/xgli/word2vec-api} for word embedding, which was trained on Google News. To cover the new words from knowledge and datasets of downstream tasks, we re-train it on knowledge SPO triples and task-specific corpus.

\subsubsection{Knowledge Graph}
We leverage HowNet\footnote{http://www\.keenage\.com/} and CN-DBpedia\footnote{http://kw.fudan.edu.cn/cndbpedia/intro/} for Chinses tasks, DBpedia\footnote{https://wiki.dbpedia.org/} and MovieKG\footnote{http://www.openkg.cn/dataset/keg-bilingual-moviekg} for English tasks. We donot do any preprocessing for these KGs, since only the matched knowledge SPO triples will be used for training K-XLNet. Detail information of the KGs are as follows.
\begin{itemize}
	\item \textbf{HowNet} is a large-scale language knowledge base for Chinese vocabulary and concepts\cite{Hownet}. It contains 52576 sets of SPO triples.
	\item \textbf{CN-DBpedia}\cite{CN-DBpedia} is a largescale 	open-domain encyclopedic KG developed by the Knowledge Work Laboratory of Fudan University, It contains 5168865 sets of SPO triples.
	\item \textbf{DBpedia} is a large-scale, multilingual knowledge base extracted from Wikipedia\cite{dbpedia}.
	\item \textbf{MovieKG} is a bilingual movie knowledge graph constructed by the knowledge engineering laboratory of the department of computer science, Tsinghua university. Unfortunately, the database is offline at present. It includes 23 concepts, 91 attributes, 0.7+ million entities and 10+ million triples. Its data sources include LinkedIMDB, baidu baike, douban, etc. 
	\item \textbf{MedicalKG} is a map of Chinese medical knowledge, which contains 13864 sets of SPO triples.
\end{itemize}

\subsubsection{Baselines}
The proposed knowledge injection method is model-independent. Therefore, we compare K-XLNet to two released XLNet models\footnote{https://github.com/zihangdai/xlnet}: \textbf{XLNet-Base} and \textbf{XLNet-Large}. Following XLNet, we design two K-XLNet models as follows: 
\begin{itemize}
	\item \textbf{K-XLNet-Base} has 3-layer, 128-hidden and 2-heads in knowledge injection layer, with the same Transformer-XL parameters as XLNet-Base.
	\item \textbf{K-XLNet-Large} has 4-layer, 128-hidden, and 3-heads in knowledge injection layer, with the same Transformer-XL parameters with XLNet-Large.
\end{itemize}

\begin{table}[h]
	\caption{Results on emotion classification task (Acc: \%)}
	\setlength{\tabcolsep}{3pt}
	\label{ect}
	\begin{center}
		\begin{tabular}{|l|rr|}
			\hline
			Method & Dev & Test \\ \hline\hline
			XLNet-Base & 95.32 & 94.37 \\
			K-XLNet-Base (DBpedia) & 95.51 & 94.88 \\
			K-XLNet-Base (MovieKG) & \textbf{95.82} & \textbf{95.03} \\ \hline\hline
			XLNet-Large & 96.21 & 95.13 \\
			K-XLNet-Large (DBpedia) & 96.74 & 95.62 \\
			K-XLNet-Large (MovieKG) & \textbf{96.87} & \textbf{95.99} \\ \hline
		\end{tabular}
	\end{center}
\end{table}

\subsection{Domain-specific task}
We first compare the performance of K-XLNet with the original XLNet on an English domain-specific task, namely emotion classification for movie reviews. 

To be specific, we use the IMDB\cite{imdb2011} dataset for this test. It includes 25,000 positive reviews and 25,000 negative reviews. We divided it into three parts: \textit{train}, \textit{dev}, and \textit{test}. We used the \textit{train} part to fine-tune the model and then evaluated its performance on the \textit{dev} and \textit{test} parts. For knowledge injection in K-XLNet, we used MovieKG and DBpedia respectively. Table~\ref{ect} shows the experimental results.       

It can be seen that K-XLNet is superior to the original XLNet in both parameter settings (base and large). This shows that our approach of knowledge injection to XLNet is effective. In addition, MovieKG performs better than DBpedia, indicating that domain knowledge is preferred for specific domain tasks.

\begin{table*}[!htbp]
	\caption{Results of various models on various open-domain tasks (Accuracy\%)}
	\setlength{\tabcolsep}{3pt}
	\label{tab2}
	\begin{center}
		\begin{tabular}{|l||cccccc||cccc||cc|}
			\hline
			\multirow{2}{1cm}{\textbf{Models$\backslash$Datasets}} & \multicolumn{2}{c}{\textbf{Book review}} & 
			\multicolumn{2}{c}{\textbf{Shopping}} &
			\multicolumn{2}{c||}{\textbf{Weibo}} &
			\multicolumn{2}{c}{\textbf{XNLI}} &
			\multicolumn{2}{c||}{\textbf{LCQMC}} &
			\multicolumn{2}{c|}{\textbf{MSRA-NER}} \\ \cline{2-13}
			& Dev & Test & Dev & Test & Dev & Test & Dev & Test & Dev & Test & Dev & Test \\ \hline\hline
			XLNet & 88.71 &	87.69 &	96.82 &	96.73 &	98.04 &	97.98 &	76.87 &	76.33 &	88.79 &	87.25 &	95.08 &	94.97 \\ \hline\hline
			K-XLNet (Hownet) & \textbf{88.83} &	\textbf{88.67} & \textbf{97.04} & \textbf{97.14} & \textbf{98.17} &	98.05 &	\textbf{77.18} & \textbf{77.13} & \textbf{89.02} & \textbf{87.37} & 96.29 & \textbf{96.26}\\	
			K-XLNet (CN-DBpedia) & 88.76 &	87.71 &	96.89 &	96.77 &	98.12 &	\textbf{98.41} &	76.97 &	76.39 &	88.87 &	87.31 &	\textbf{96.31} & 96.24 \\\hline
		\end{tabular}
	\end{center}
\end{table*}

\begin{figure}[h]
	\vspace{-6mm}
	\centering
	\includegraphics[width=8.5cm,height=5cm]{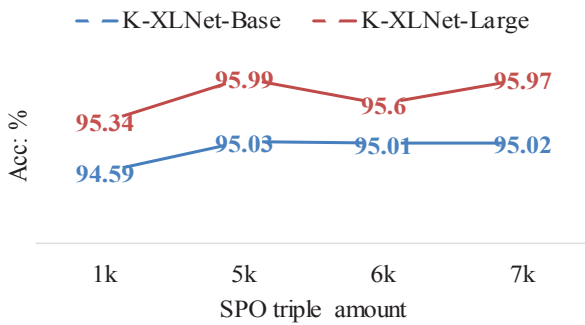}
	\caption{Performance of K-XLNet models injected different triple (knowledge) amounts.}
	\label{fig:kga}
\end{figure}

We further investigate the effect of different SPO triple (knowledge) amounts in K-XLNet. In this test, we use MovieKG for knowledge injection and set the amount of knowledge triples to be 1,000, 5,000, 6,000, and 7,000 respectively. The results are shown in Figure~\ref{fig:kga}.

We can see that from 1K to 5K, the performances of K-XLNet models are improving with the increase of knowledge injection. After Performance that, the performance tends to be stable or even slightly decreased. This gives us a hint that when using knowledge for model improvement, more is not better. 

In the following experiments, we set the triple (knowledge) amount to 5k for all K-XLNet models, and compare our K-XLNet to the original XLNet using the \textit{Large} setting, since the \textit{Base} setting has similar performance trend.

\subsection{Open domain tasks}
We conduct six experiments to evaluate the performance of K-XLNet on open-domain tasks. Specifically, \textbf{Book\_review}\cite{liu2019k}, \textbf{Shopping}\cite{liu2019k} and \textbf{Weibo}\cite{liu2019k} are single-sentence classification tasks. \textbf{XNLI}\cite{XNLI18} and \textbf{LCQMC}\cite{LCQMC18} are two-sentence classification tasks. \textbf{MSRA-NER}\cite{Levow06} is a Named Entity Recognition (NER) task.

\begin{itemize}
	\item \textbf{Book\_review}\footnote{https://embedding\.github\.io/evaluation} contains 20,000 positive reviews and 20,000 negative reviews collected from \href{https://book.douban.com/}{Douban}.
	\item \textbf{Shopping}\footnote{https://share\.weiyun\.com/5xxYiig} is a online shopping review dataset
	that contains 40,000 reviews, including 21,111 positive reviews and 18,889 negative reviews.
	\item \textbf{Weibo}\footnote{https://share.weiyun.com/5lEsv0w} is a dataset with emotional annotations from Sina Weibo, including 60,000 positive samples and 60,000 negative samples.
	\item \textbf{XNLI}\footnote{https://github.com/NLPchina/XNLI} is a cross-language language understanding dataset in which each entry contains two sentences and the task is to determine their relation (``Entailment'', ``Contradict'' or ``Neutral''). 
	\item \textbf{LCQMC}\footnote{https://github.com/Lizhengo/lcqmc\_data} is a large-scale Chinese question matching corpus. The goal of this task is to determine if the two questions have a similar intent.
	\item \textbf{MSRA-NER}\footnote{https://github.com/littleWhiteTre/msra\_ner/} is a NER dataset published by Microsoft. It is to recognize the named entities in the text, including person names, place names, organization names, etc..	
\end{itemize}

Similarly, the open-domain datasets are split into three parts: \textit{train}, \textit{dev}, and \textit{test}, used for fine-tuning, model selection and model test, respectively. Table~\ref{tab2} shows the test results of various models in terms of accuracy. We can see that K-XLNet performs better than XLNet consistently on the six open-domain tasks. To be specific, the improvements are significant on NER task, but not on sentence classification tasks. Moreover, the model leveraging Hownet performs better than that using Cn-DBpedia on sentence classification tasks, but It is the opposite on the NER task on the NER task. The above observations show that knowledge injection to XLNet is also effective on open-domain downstream tasks, but it is important to choose appropriate knowledge base according to the specific task.

\section{Conclusion}
\label{sec:Conclusion}
In this paper, we propose a simple but general knowledge injection method for pretraining language model. Taking the XLNet as a running example, we construct K-XLNet to show the effectiveness of the proposed method. Extensive experiments show that K-XLNet performs better than XLNet in both open domain and domain-specific tasks, and the improvement on domain-specific task is more significant than that on open domain task. However, there is still much room for improvement such as improving the interpretability of the sentence generator, reducing the impact of knowledge noise, and optimizing the role of structured information of knowledge.

\bibliography{KXLNet.bib}
\bibliographystyle{IEEEtran}

\newpage
\begin{IEEEbiography}[{\includegraphics[width=1in,height=1.25in,clip,keepaspectratio]{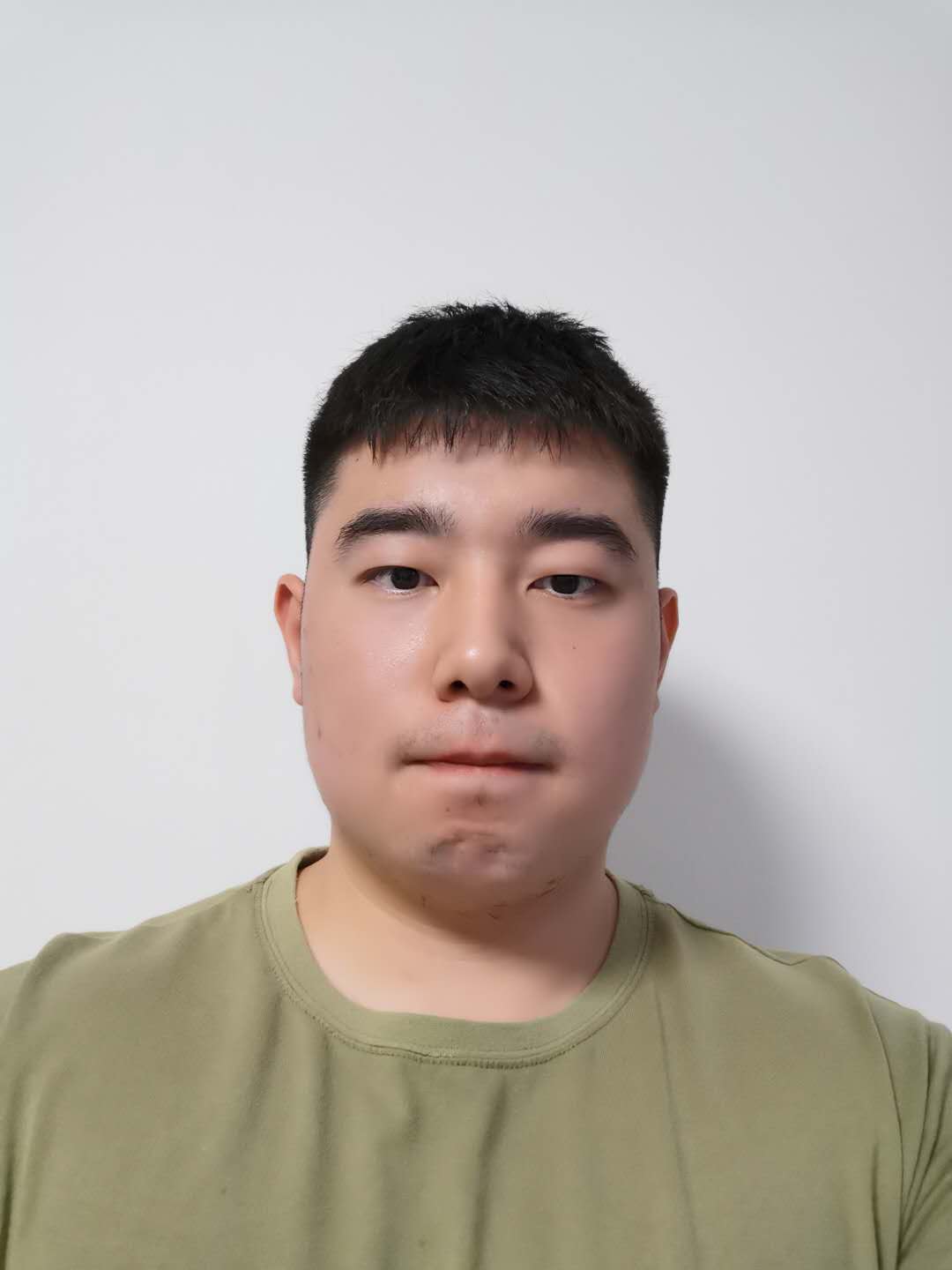}}]{Ruiqing Yan} is currently a junior at Beijing institute of petrochemical technology in Beijing, China. Majoring in computer science and technolegy, He is very interested in deep learning technology and its applications in naturl language processing.
\end{IEEEbiography}

\begin{IEEEbiography}[{\includegraphics[width=1in,height=1.25in,clip,keepaspectratio]{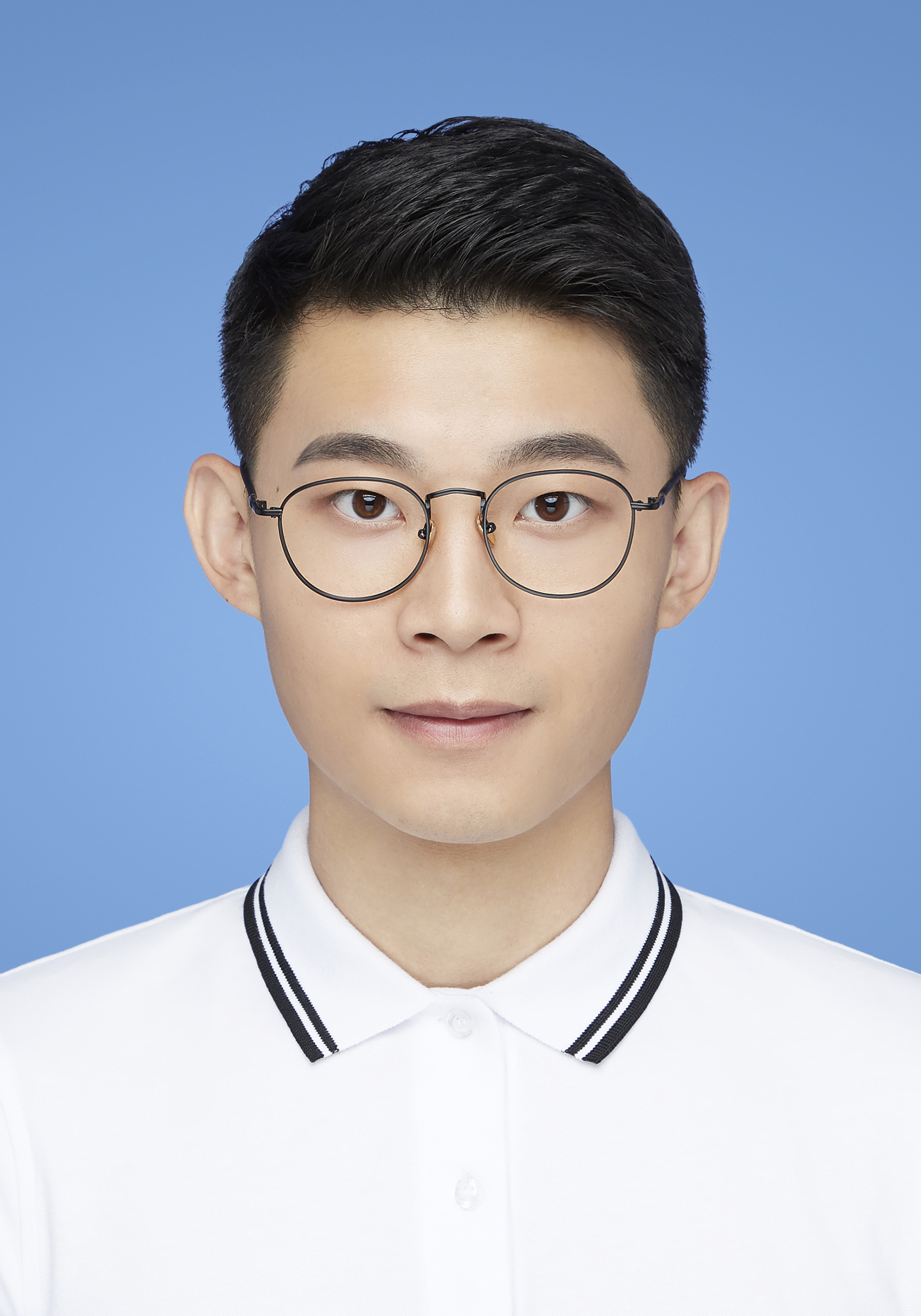}}]{Lanchang Sun} received his B.S. degree in computer science from Beijing institute of petrochemical technology in 2018. He is currently pursuing a master's degree at Beijing University of Posts and telecommunications in Beijing, China. His research interests include machine learning and pattern recognition, especially the application of deep learning techniques in natural language processing and image recognition.
\end{IEEEbiography}

\begin{IEEEbiography}[{\includegraphics[width=1in,height=1.25in,clip,keepaspectratio]{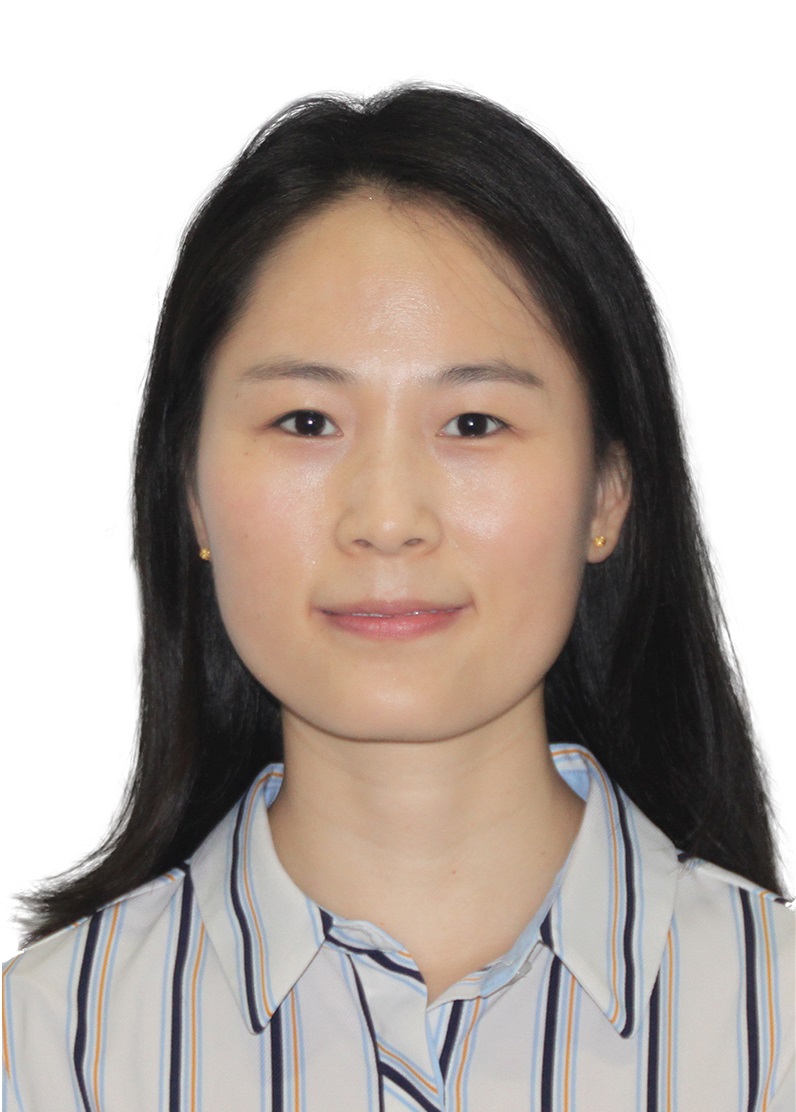}}]{Fang Wang} graduated with B.S. degree and M.S. degree in computer science from Beijing Technology and Business Universtiy in 2008 and 2011, respectively. She received her Ph.D. degree in Computer Science from Beihang University in 2017. After that, she works with Beijing Institute of Petrochemical Technology. Her research interests include data mining, knowledge engineering and natural language processing.
\end{IEEEbiography}

\begin{IEEEbiography}[{\includegraphics[width=1in,height=1.25in,clip,keepaspectratio]{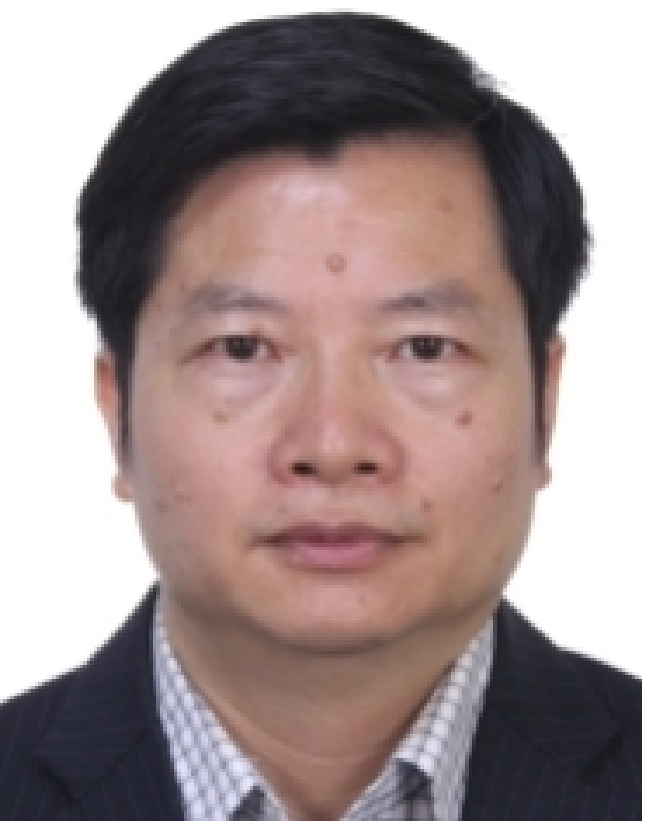}}]{Xiaoming Zhang} is working as a professor of Beijing Institute of Petrochemical Technology. He graduated with B.S. degree from Zhejiang University in 1989, received his M.S. and Ph.D. degree in Automation Engineering from Dalian University of Technology in 1992 and 1996 respectively. His current research interests include data mining, machine learning and big data technology.
\end{IEEEbiography}
\EOD
\end{document}